\documentclass[sigconf]{acmart}

\usepackage{amsthm}
\theoremstyle{definition}

\usepackage{algorithm,algorithmic,amsmath}
\newcommand{\squeezeup}{\vspace{-2.5mm}} 
\theoremstyle{remark}

\usepackage{enumitem}
\setlist[enumerate,1]{label=(\roman*)}
\setlist[enumerate,2]{label=(\alph*)}

\def\BibTeX{{\rm B\kern-.05em{\sc i\kern-.025em b}\kern-.08emT\kern-.1667em\lower.7ex\hbox{E}\kern-.125emX}}

\copyrightyear{2019}
\acmYear{2019} 
\acmConference[RecSys Challenge '19 Companion]{Companion Proceedings of the 2019 RecSys Conference}{May 16--20, 2019}{Copenhagen, Denmark}
\acmBooktitle{}
\acmPrice{}
\acmDOI{}
\acmISBN{}

\begin{document}

\title{Many-to-one Recurrent Neural Network for \\ Session-based Recommendation}

\author{Amine Dadoun}
\affiliation{EURECOM, Sophia Antipolis, France }
\affiliation{Amadeus SAS, Biot, France}
\email{amine.dadoun@eurecom.fr}

\author{Rapha{\"{e}}l Troncy}
\affiliation{EURECOM, Sophia Antipolis, France }
\email{raphael.troncy@eurecom.fr}

\begin{abstract}
This paper presents the D2KLab team's approach to the RecSys Challenge 2019 which focuses on the task of recommending accommodations based on user sessions. What is the feeling of a person who says ``Rooms of the hotel are enormous, staff are friendly and efficient''? It is positive. Similarly to the sequence of words in a sentence where one can affirm what the feeling is, analysing a sequence of actions performed by a user in a website can lead to predict what will be the item the user will add to his basket at the end of the shopping session. We propose to use a many-to-one recurrent neural network that learns the probability that a user will click on an accommodation based on the sequence of actions he has performed during his browsing session. More specifically, we combine a rule-based algorithm with a Gated Recurrent Unit RNN in order to sort the list of accommodations that is shown to the user. We optimized the RNN on a validation set, tuning the hyper-parameters such as the learning rate, the batch-size and the accommodation embedding size. This analogy with the sentiment analysis task gives promising results. However, it is computationally demanding in the training phase and it needs to be further tuned.
\end{abstract}



\maketitle

\section{Introduction}
\label{sec:introduction}
With the growing desire to travel on a day-to-day basis, the number of accommodation offers a user can find on the Web is increasing significantly. It has therefore become important to help travellers choosing the right accommodation to stay among the multitude of available choices. Customizing the offer can lead to a better conversion of the offers presented to the user. Recommender Systems play an important role in order to filter-out the undesired content and keep only the content that a user might like. In particular, in a user's navigation settings, session-based recommender systems can help the user to more easily find the elements she wants based on the actions she has performed. Most popular recommender system approaches are based on historical interactions of the user. This user's history allows to build a long-term user profile. However, in such setting, users are not always known and identified, and we do not necessarily have long-term user profile for all users. Traditional models propose to use item nearest neighbor schemes to overcome this user cold start problem \cite{Linden03} or association rules in order to capture the frequency of two co-occurring events in the same session \cite{Agrawal93}. In recent years, some research works have focused on recurrent neural networks (RNNs) \cite{hidasi15} considering the sequence of user's actions as input of the RNN. The RNN learns to predict the next action given a sequence of actions.

\begin{figure}[h]
    \includegraphics[width=110mm]{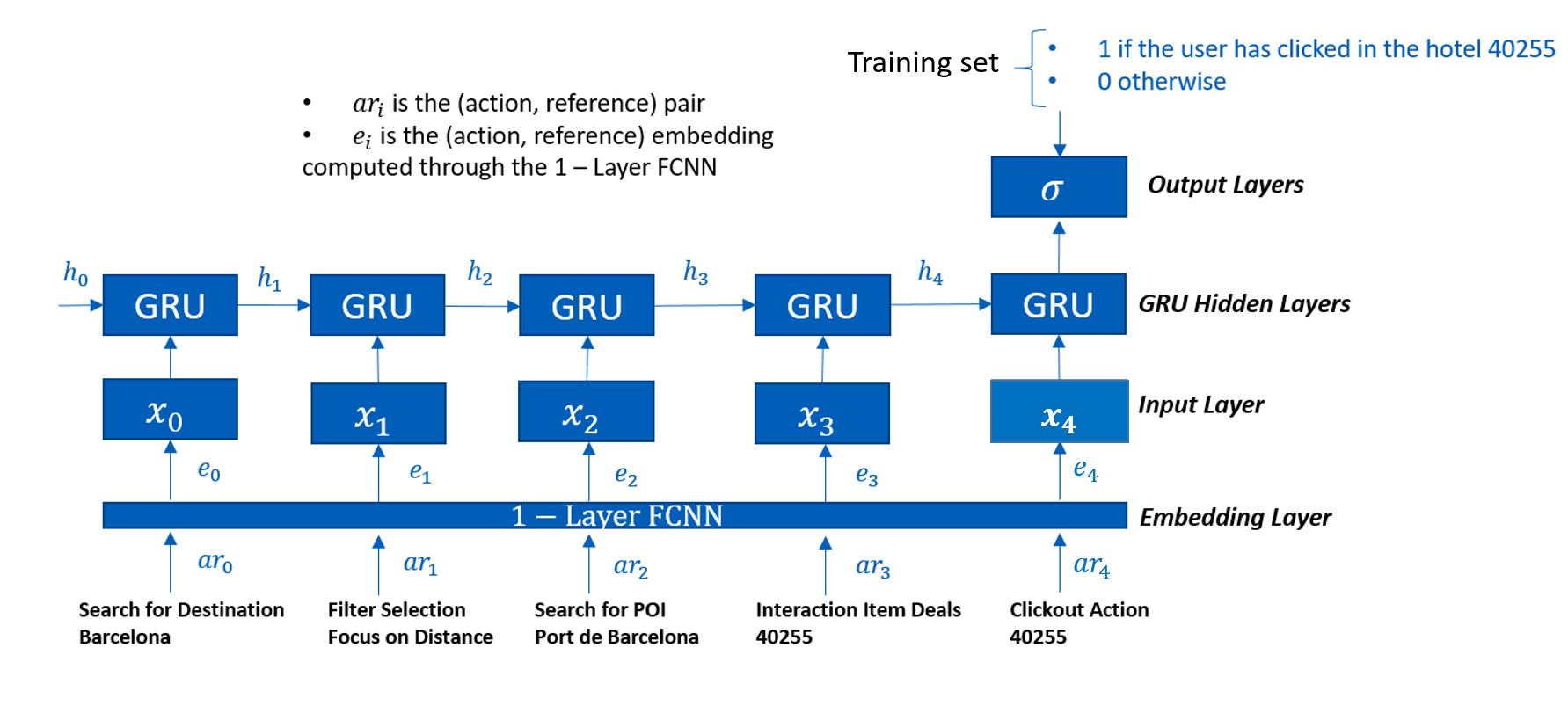}
    \caption{Many-to-one Recurrent Neural Network}
    \label{fig:Many_to_one_rnn}
\end{figure}

Inspired by this recent work, we propose a many-to-one recurrent neural network that predicts whether or not the last item in a sequence of actions is observed or not as illustrated in figure~\ref{fig:Many_to_one_rnn}. More formally, our RNN returns the probability that a user clicks on an item given the previous actions made in the same session: $P(r_{t}|ar_{t-1}, ar_{t-2}, ..., ar_{0})$, where $r_{t}$ is the item referenced by $40225$ in this example (see figure~\ref{fig:Many_to_one_rnn}). This value is then combined with a rule-based approach that explicitly places the elements seen in the previous steps at the top of the accommodation list displayed to the user. To the best of our knowledge, this is the first time where many-to-one RNN architecture is used for session-based recommendation, and this represents our main contribution.

The remainder of the paper is organized as follows. Section~\ref{sec:dataset} provides an exploratory data analysis of the Trivago dataset\footnote{\url{http://www.recsyschallenge.com/2019/}}. In Section~\ref{sec:approach}, we present the approach to build the many-to-one recurrent neural network and combine it with the rule-based approach. Section~\ref{sec:experiments} presents the experiments carried out to show the effectiveness of our model and an empirical comparison with some session-based recommender baseline methods. Finally, in Section~\ref{sec:conclusion}, we provide some conclusions and we discuss future works.

\section{Dataset}
\label{sec:dataset}
The first part of our work consists of conducting an exploratory data analysis to understand user behavior on the Trivago website\footnote{\url{https://www.trivago.com}} and then interpret the results we will obtain from our models. The dataset published for the challenge consists of interactions of users browsing the trivago website collected from 01-11-2018 to 09-11-2018 (9 days). More precisely, for a given session, we have the sequence of actions performed by the user, the filters applied, the accommodation list displayed to the user when performing a clickout action, plus the price of each accommodation in the list. In addition to this information, we have two contextual features: the device and the platform used by the user to perform the searches. The remainder of this section presents statistics and overviews of training data.

\begin{figure}[h]
    \includegraphics[width=\linewidth]{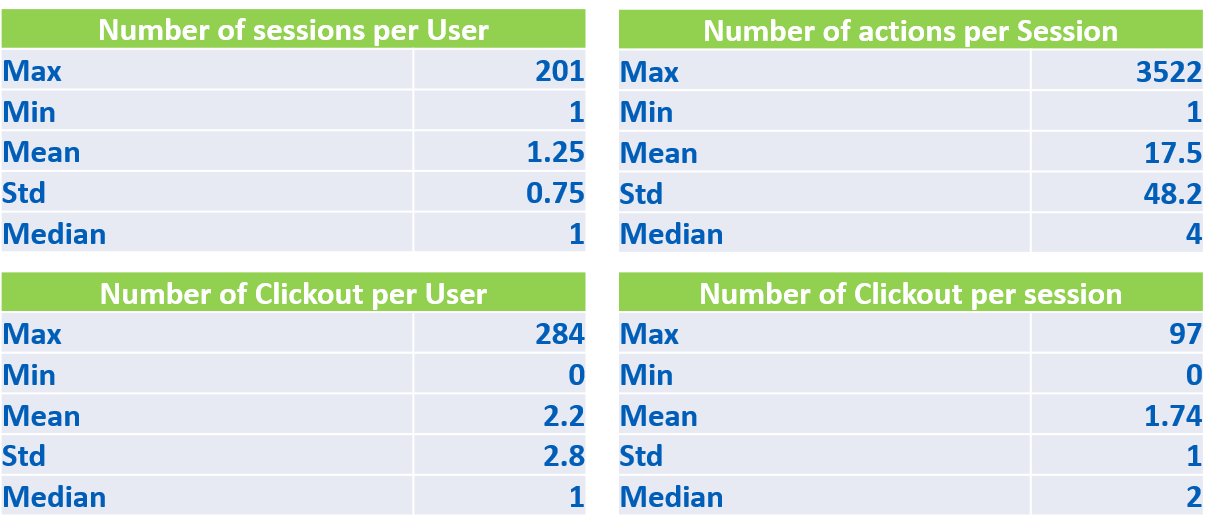}
    \caption{Statistics on Trivago dataset}
    \label{fig:Stats_Trivago}
\end{figure}

\textbf{General statistics on training data:} We report in Figure~\ref{fig:Stats_Trivago} 5 summary statistics of different variables that characterize user sessions. The statistic tables highlight two important observations:
\begin{itemize}
    \item \textit{Dispersion}: The number of actions per session has a high standard deviation which means that the data is highly spread. For all the variables, we also have a very high maximum value which demonstrates the skewness of users' behaviors.
    \item \textit{Actions required for 'Clickout Action'}: On average, a user performs $17.5$ actions in a session. However, the average number of actions needed to perform a 'Clickout Action' is only $8$, so what does the rest of the clicks correspond to? In more than $72 \%$ of cases, the last performed action in a session is a 'Clickout Action'. However, in $28 \%$ of all sessions, there are other actions following the click out action.
\end{itemize}

\squeezeup
\squeezeup

\textbf{Filters and sort by actions:} One could also be interested to know more about some variables' distributions. We have plotted the histogram of most of the 15 filters used. The observed distribution does not follow a long-tail distribution and all filters are more or less used in similar proportion. We can thus infer that there are different types of user behaviors. More specifically, we compute the ratio of sessions where users use filter or sort buttons: this ratio is equal to $14 \%$ which represents a significant subset of the data. We also compute the average number of clickout actions performed per session for each platform and noticed that there is a significant difference between people that are searching for accommodation using the Japan platform ($8.7$ clickout actions) and the Brazil one ($23.9$ clickout actions). Finally, we compute the average time a user spends in a session ($8$ minutes), and we noticed that there is a high standard deviation for this variable ($22$ minutes) which again demonstrates the dispersion in users' sessions. This leads us to the conclusion that there are different user profiles and behaviors. For example, we have users who need a lot of actions to finally perform a clikckout, users who perform volatile clicks, users who have to look at the images of the accommodation and then click on it, etc. Explicitly adding this information to our model can help to more effectively predict the user's clickout element. In addition, the idea of having a different model for each type of user is something that should be experimented with.


\textbf{Accommodation Content:} In addition to information on user sessions, a description of the accommodation is also provided. This enriches the input data of our recommendation system. We have $157$ different properties that describe an accommodation (wifi, good rating, etc.). We use these properties to enrich the input data of the RNN as explained in~\ref{fig:acc_cnt}.


\section{Approach}
\label{sec:approach}
\begin{figure}[h]
    \includegraphics[width=100mm,scale=0.5]{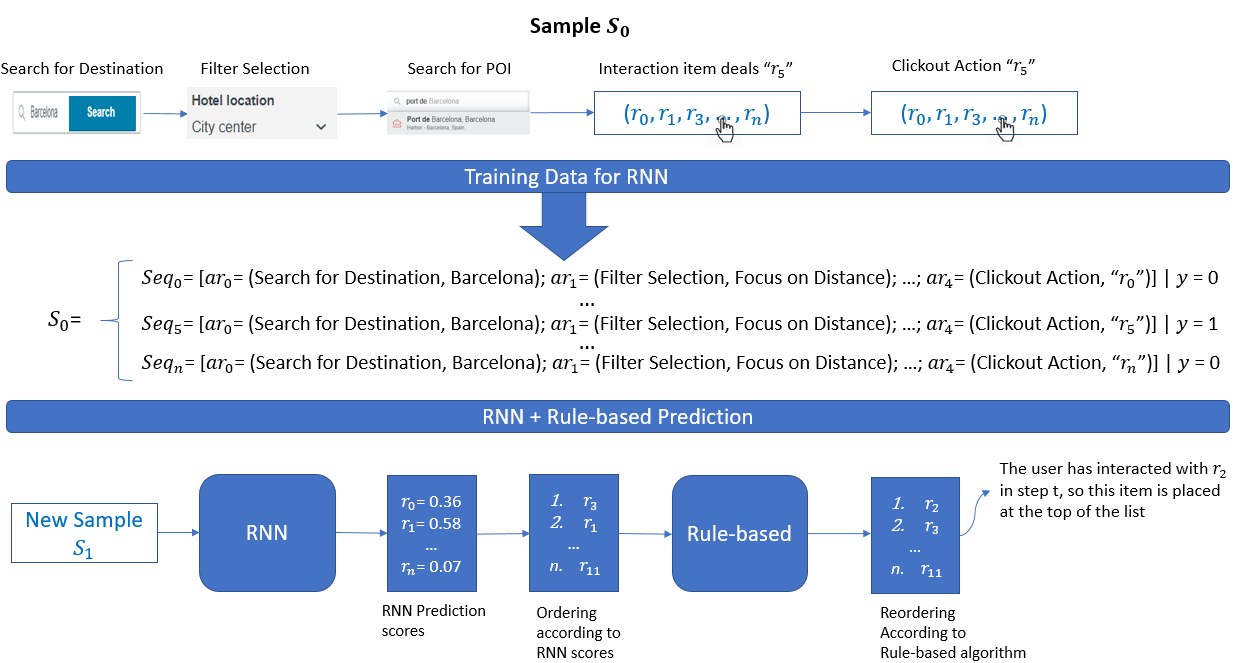}
    \caption{Our approach}
    \label{fig:acc_cnt}
\end{figure}

Our approach is a two-stage model that consists in computing a score for each element of the impression list displayed to the user when she performs a clickout, based on the RNN, and then applying a rule-based algorithm to the ranked list returned by the RNN (Figure~\ref{fig:acc_cnt}). 

\squeezeup

\subsection{Recurrent Neural Network}
\label{subsec:RNN}
Recurrent neural networks are widely used for many NLP tasks such as named entity recognition, machine translation or semantic classification \cite{TaiSM15}. Indeed, this architecture works very well when it comes to recognizing sequence-based patterns and predicting the following element from a sequence of previous elements. It was therefore natural to use this neural network architecture to predict the next click based on the sequence of actions performed by the user. However, unlike \cite{Sutskever11}, we consider our problem as a binary classification instead of a multi-label classification problem. More precisely, the RNN takes as input a sequence of actions with their corresponding references, represented by a one-hot encoding vector and fed into a one fully connected neural network in order to compute the (action, reference) embeddings, plus the last action that corresponds to a clickout with its reference, and then returns $P(r_{t}|ar_{t-1}, ar_{t-2}, ..., ar_{0})$, where $ar_{i}$ indicates the (action, reference) pair made by the user at step $i$. This probability indicates if the user has clicked in the accommodation $r_{t}$ given the sequence of previous (action, reference) pairs $(ar_{t-1}, ar_{t-2}, ..., ar_{0})$. Therefore, for each clickout action, our RNN returns a score for each item in the accommodation list, and the list of accommodations is reorganized in a decreasing way according to the score of the items.

\begin{figure}[h]
    \includegraphics[width=\linewidth]{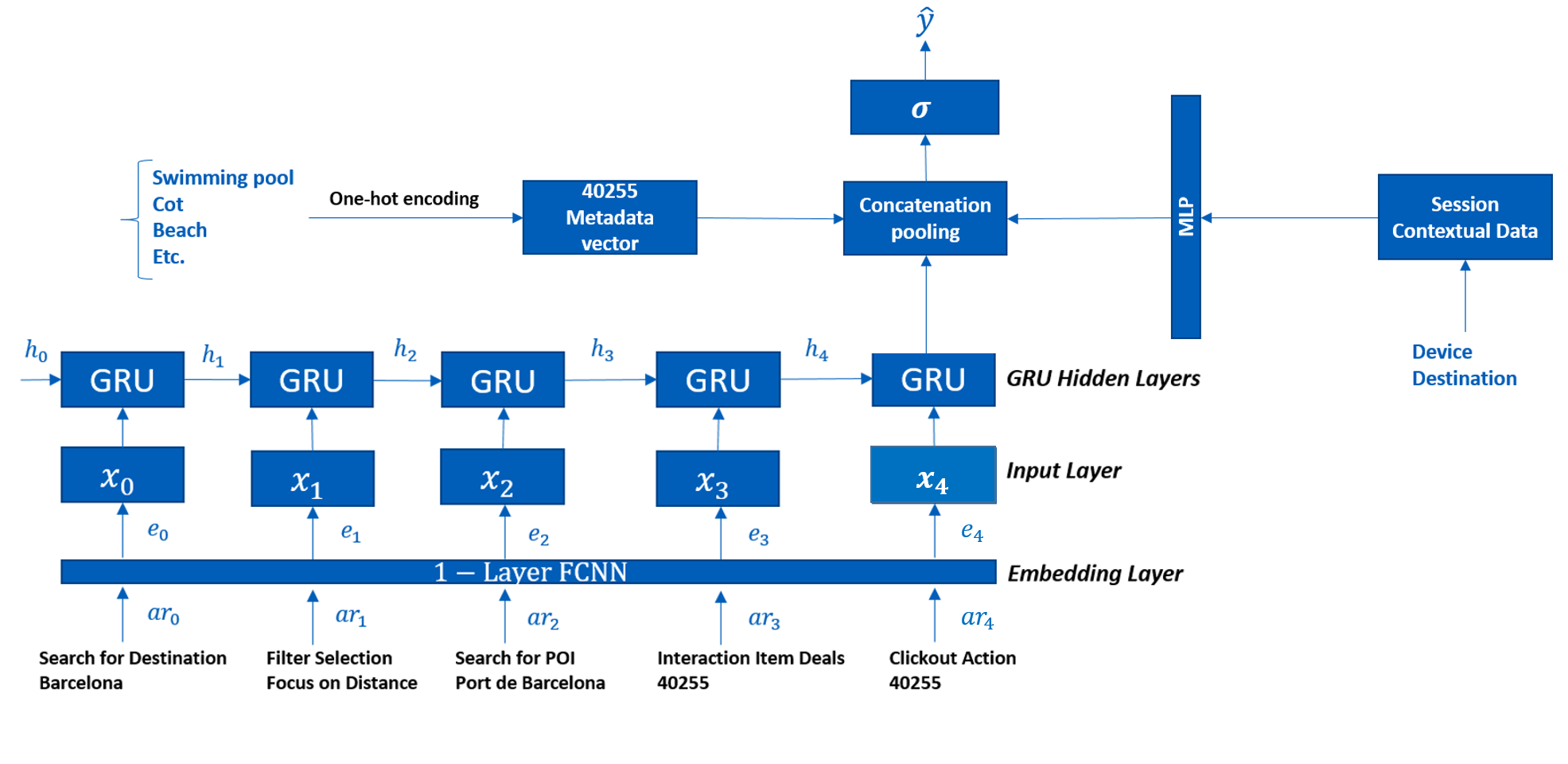}
    \caption{RNN and MLP combination for content and context information}
    \label{fig:rnn_2}
\end{figure}

In addition to the sequence of actions performed by the user, we first enrich our input data with the content of the accommodation, and then we add the contextual information of the session as shown in figure~\ref{fig:rnn_2}. The content information are represented using one hot-encoding technique where each element of the vector corresponds to a property (e.g. wifi, restaurant, etc.) that represents the accommodation. We use the device and the platform as session-contextual information. These two categorical features are one-hot encoded as well and fed into a Multi-layer perceptron (MLP) as represented in Figure~\ref{fig:rnn_2}. The MLP used is a 2 layer feed-forward neural network. The size of each layer is being optimized using Grid search as specified in Section~\ref{subsec:results}. We use GRU cells \cite{ChungGCB14} in order to compute the hidden states $h_{t}$ for each step $t$ and a sigmoid function in order to compute the probability score $\hat{y} = \sigma(W_{y} h_{t} + b_{y})$, where $\sigma(x) = \frac{1}{1+e^{-x}}$. 

\squeezeup
\squeezeup

\subsection{Rule-based Algorithm}
\label{subsec:rule-based algo}
Contrarily to the RNN algorithm which predicts a score for each element in the accommodation list, the rule-based algorithm simply reorders the accommodation list based on explicit prior items the user interacted with in previous actions. The motivation to use this rule-based algorithm was the analysis made beforehand on the data which showed us an interesting and recurrent pattern: in several sessions, users who have interacted with an accommodation have performed a clickout action on this accommodation slightly later, where interacting with an accommodation is among the following actions: \{Interaction item ratings, Interaction item deals, Interaction item image, Interaction item information, Search for item, Clickout item\}.
More precisely, we consider, among the elements in the accommodation list, those that were interacted with before the clickout action, we name this ensemble of elements $I_{acc}$. The closer the element in $I_{acc}$ is to the clickout action, the higher it is placed at the top of the list as illustrated in the Figure~\ref{fig:acc_cnt}.

\section{Experiments}
\label{sec:experiments}
Trivago published users' session data split into training and test set. The training set has been fully used in order to train our RNN. The test set, as proposed by the organizers\footnote{\url{http://www.recsyschallenge.com/2019/Dataset}}, has been split into validation set in order to compute scores of our model in local and confirmation set which is the subset of the data used to submit the results in the submission page\footnote{\url{http://www.recsyschallenge.com/2019/submission}}. 
As proposed by the organizers, we have used mean reciprocal rank as metric to evaluate the algorithm, which can be defined as follows:
    \begin{equation}
        MRR = \frac{1}{n} \sum_{t=1}^{n} \frac{1}{rank(c_{t})},
    \end{equation}
where $c_{t}$ represents the accommodation that the user clicked on in the session $t$.

We also implement a set of baseline methods (Section~\ref{sec:baselines}) with which we compare our method.

\subsection{Baseline Methods}
\label{sec:baselines}
Most used recommender systems are based on a long-term user history which lead one to implement methods such as matrix factorization \cite{Koren09} as a baseline. However, in session-based recommender system, we do not have such long user past interaction \cite{Ludewig18}. Different baselines were implemented in the setting of session-based recommendation as proposed in \cite{Ludewig18}, and are described bellow:
\begin{itemize}
    \item Association rules \cite{Agrawal93}: The association rule is designed to capture the frequency of two co-occurring events in the same session. The output of this recommender system is to give a ranking of next items based on a current item.
    \item Markov Chains \cite{norris97}: Similarly to the association rules approach, markov chains is also capturing the co-occurring events in the same session, but only takes into account two events that follow one after the other(in the same session).
    \item Sequential rules \cite{Kamehkhosh17}: This method is similar to association rules and markov chains since it tries to capture frequency of co-occurring events, but it adds a weight term that captures the distance between the two occurring events.
    \item IKNN \cite{hidasi15}: Each item is represented by a sparse vector $V_{i}$ of length equal to the number of sessions, where $V_{i}[j] = 1$ if the item is seen in the session j and $0$ otherwise. 
\end{itemize}

\subsection{Experimental Results}
\label{subsec:results}
\textbf{Implementation Framework \& Parameter Settings:} 
Our model and all the baselines are implemented using Python and Tensorflow library\footnote{Python Tensorflow API: \url{https://www.tensorflow.org}}. The hyper-parameters of the RNN were tuned using grid-search algorithm. First, we initialized all the weights randomly with a Gaussian Distribution ($\mu = 0$, $\sigma = 0.01$), and we used mini-batch Adam optimizer \cite{Kingma14}. It is worth mentioning that other optimizers were tested. However, Adam Optimizer has shown to be the most efficient in time and also accuracy. We evaluated our model using different values for the hyper-parameters: 
\begin{itemize}
    \item Size of $ar$ embeddings: $E\_size \in \{64, 128, 256, 512\}$
    \item Hidden state: $h\_size \in \{64, 128\}$
    \item Batch size: $B\_size \in \{32, 64, 128, 256, 512\}$
    \item Number of epochs: $epochs \in \{5, 10, 15, 20\}$
    \item Learning rate: $l_{r} \in \{0.0001, 0.0005, 0.001, 0.005, 0.01\}$
    \item MLP layers size: $l\_sizes \in \{[256, 128], [128, 64], [64, 32]\}$
\end{itemize}

\textbf{Results \& Discussion:}
The results are reported in table~\ref{tab:results}. The scores correspond to an average of numerous experiments of the MRR metric computed on the validation set proposed by the organizers. The rule-based approach is the most effective with a score of $0.56$. The ranking of these methods remains the same for the confirmation set, where the best score was obtained using the rule-based algorithm ($MRR = 0.648$). The RNN does not work as expected, even when session context information and accommodation meta-data have been added to the model. Association rules and sequential rules give promising results: $0.52$ and $0.51$ respectively, when the Markov chains give only a score of $0.34$. This shows that it is more important to consider all the elements seen in a session as close to each other than to consider only those seen sequentially close to each other. Finally, our approach of combining the RNN and the rule-based method is less successful than the simple rule-based method. This is because the order obtained by the RNN is worse than the order given by Trivago for the rule-based method, even if the order given by the RNN has better raw results than the order given by Trivago when calculating the MRR.

\begin{table}
  \caption{MRR scores on Validation set}
  \label{tab:results}
  \begin{tabular}{cl}
    \toprule
    Model& MRR\\
    \midrule
    Association Rules & 0.52 \\
    Markov Chains & 0.34\\
    Sequential Rules & 0.51\\
    IKNN & 0.54\\
    RNN & 0.49\\
    RNN + Metadata & 0.50 \\
    RNN + Context & 0.49 \\
    RNN + Metadata + Context & 0.50 \\
    \textbf{Rule-based} & 0.56 \\
    RNN + Metadata + Context + Rule-based & 0.54 \\
  \bottomrule
\end{tabular}
\end{table}
\squeezeup
\subsection{Lessons Learned}
\label{subsec:lessons}
The task of predicting which element the user will click on based on performed actions has been treated in a similar way to predicting the next word in a sentence. However, while the context in a sentence is very important and plays a big role in considering that two consecutive words have a sense, hence the use of RNN, we cannot be sure that the context is just as important for our task, especially when we look at the volatility of actions made by users in the same session. This leads us to question ourselves, especially when we look at the results obtained from the association rules and the method of the K-nearest neighbors which are better than the Markov chain method or the sequential rule method. This demonstrates that the succession of actions is not as important as it is assumed at the beginning of our study, and that the simple fact of considering the set of actions than the sequence of actions could have been better. 

The second important point to emphasize is the dispersion of user behavior with the website: indeed, when analyzing the data, we noticed that there are several types of users, which makes it complicated and difficult to build a model for all types of users and to find a pattern that generalizes all the different behaviors in order to make accurate predictions for our task. The idea proposed during the data analysis which is to create different models per user seems to be a good idea as well. 

Lastly, the simple rule-based method is the most efficient and is not as far from the method that obtained the best result in this challenge (0.648 against 0.689). Given that this method does not require any learning, nor much computation time, it is worth using this method for obvious use cases as the example shown in Figure~\ref{fig:acc_cnt}.  

\squeezeup

\section{Conclusion and Future Work}
\label{sec:conclusion}
Recommender systems help users to find relevant elements of interest to facilitate navigation and thus increase the conversion rate in e-commerce sites or build user loyalty in streaming video sites. In this challenge, the aim is to help the user to easily find the accommodation in which she wants to go, and to place it in the top of a list of different accommodations that are proposed to her, given previous performed actions in a session. However, by having only the actions performed by the user in addition to some information related to the context of the session such as the user's device or Trivago's platform, such a task is hard. Especially, in the world of travel where the context is very important, such as the season in which a user travels but also if this user travels alone, in a group or with a family. In this paper, we presented the D2KLAB solution that implements a two-stage model composed by a one-to-many recurrent neural network that predicts a score for each item in the accommodation list to reorder the list given as input and then apply a rule-based algorithm to reorder once again this list based on explicit prior action performed in the session. The implementation of our method is publicly available at \url{https://gitlab.eurecom.fr/dadoun/hotel_recommendation}.

One major outstanding question remains regarding the applicability of the methods designed for this challenge: once a user has interacted with an element of the accommodation list and has performed a clickout action on an element of this list which redirects her to the merchant website, it is already too late to change the order of this list. Furthermore, having removed the three features (namely, stay duration, number of travelers plus the arrival date) from the dataset makes those models not representative of the real traffic. In future work, we would like to exploit session actions as a set of actions instead of a sequence of actions and use an algorithm that combines different types of inputs, namely, collaboration, knowledge and contextual information in a learning model as proposed in \cite{Dadoun19}. We also envision to better segment users and to build a model per user segment.

\newpage
\bibliographystyle{ACM-Reference-Format}
\bibliography{bibliography}

\end{document}